\documentclass[runningheads]{llncs}
\usepackage{enumitem}
\usepackage{microtype}
\usepackage{mathptmx}

\usepackage{tikz}
\usetikzlibrary{positioning, shapes.geometric, fit}
\usepackage[T1]{fontenc}
\usepackage{pgfplots}
\pgfplotsset{compat=1.18}

\usepackage{xcolor}

\definecolor{barblue}{RGB}{31,119,180}
\definecolor{barsteel}{RGB}{100,160,210}
\definecolor{interbar}{RGB}{52,120,170}

\definecolor{barblue}{RGB}{31,119,180}
\definecolor{barsteel}{RGB}{100,160,210}
\definecolor{barslate}{RGB}{70,130,160}
\definecolor{barmid}{RGB}{44,160,130}
\definecolor{barslight}{RGB}{90,180,160}
\definecolor{bargray}{RGB}{150,180,195}
\usepackage{graphicx}
\usepackage{threeparttable}
\usepackage{multirow}

\usepackage{tabularx}
\usepackage{booktabs}
\usepackage{amsmath}

\begin{document}
\title{Human-in-the-Loop Benchmarking of Heterogeneous LLMs for Automated Competency Assessment in Secondary Level Mathematics}
\titlerunning{Human-in-the-Loop LLM Benchmarking}

\author{
Jatin Bhusal\inst{1} \and
Nancy Mahatha\inst{2} \and
Aayush Acharya\inst{3} \and
Raunak Regmi\inst{4}
}

\institute{
Research and Incubation Center (RAIN), Sunway College Kathmandu\\
\email{jatin@sunway.edu.np, nancymahatha@gmail.com, aayushacharya8018@gmail.com, raunak.regmi@gmail.com}
}
\maketitle              
\begin{abstract}
As Competency-Based Education (CBE) is gaining traction around the world, the shift from marks-based assessment to qualitative competency mapping is a manual challenge for educators. This paper tackles the bottleneck issue by suggesting a ``Human-in-the-Loop'' benchmarking framework to assess the effectiveness of multiple LLMs in automating secondary-level mathematics assessment. Based on the Grade 10 Optional Mathematics curriculum in Nepal, we created a multi-dimensional rubric for four topics and four cross-cutting competencies: Comprehension, Knowledge, Operational Fluency, and Behavior and Correlation.

The multi-provider ensemble, consisted of open-weight models—Eagle (Llama 3.1-8B) and Orion (Llama 3.3-70B) and proprietary frontier models Nova (Gemini 2.5 Flash) and Lyra (Gemini 3 Pro), was benchmarked against a ground truth defined by two senior mathematics faculty members ($\kappa_w = 0.8652$). The findings show a marked ``Architecture-compatibility gap''. Although the Gemini-based Mixture-of-Experts (Sparse MoE) models achieved ``Fair Agreement'' ($\kappa_w \approx 0.38$), the larger Orion (70B) model exhibited `No Agreement' ($\kappa_w = -0.0261$), suggesting that architectural compliance with instruction constraints outweighs the scale of raw parameters in rubric-constrained tasks. We conclude that while LLMs are not yet suitable for autonomous certification, they provide high-value assistive support for preliminary evidence extraction within a ``Human-in-the-Loop'' framework.

\end{abstract}

\section{Introduction}
Assessing students' performance has been a crucial part of academics historically. Traditionally, subjects such as mathematics were assessed based on students' marks and scores instead of competencies. This displays a significant gap in the assessment system, as marks alone might not be the best metric to measure student understanding of specific topics. Competencies are defined as skills and knowledge areas required to demonstrate proficiency in a subject. A better example can be: to excel in AI, a scholar needs to have knowledge in domains like: theory, mathematics, applications and ethics. These form four core competencies in AI.  A clear assessment of a student’s level in the competencies required to excel in the subject provides a more scientific, rich, and tangible analysis of where a student’s strengths lie and where weaknesses exist. This is where Competency-Based Education (CBE) plays a crucial role in addressing limitations of traditional assessment practices \cite{chappell2020competency}. 

With the current reasoning capacity of Large Language Models (LLMs), free open-source versions and proprietary models can be leveraged to assess student performance beyond the quantitative window, including qualitative aspects, analogous to human supervision. LLMs can reason over open-ended answers; however, no accurate comparison of LLMs vs. expert teachers has yet been conducted at the competency-rubric level, especially in school mathematics. The Global South’s academic diaspora is particularly behind in this aspect, with many remaining gaps in integrating modern academic practices and bridging the digital divide \cite{leach2006deep}. 

\begin{figure}[t]
\centering
\resizebox{\linewidth}{!}{
\begin{tikzpicture}[
    node distance=0.6cm,
    every node/.style={font=\sffamily},
    box/.style={draw, rounded corners, align=center, minimum width=2.8cm, minimum height=1.2cm, thick},
    arrow/.style={->, >=stealth, very thick, gray!80}
]

\node[box, fill=blue!10, draw=blue!50] (step1) {\textbf{1. Rubric Construction}\\ \footnotesize Expert-designed\\ \footnotesize competency framework};
\node[box, fill=blue!10, draw=blue!50, right=of step1] (step2) {\textbf{2. Assessment Design}\\ \footnotesize 16 open-ended\\ \footnotesize math questions};
\node[box, fill=orange!10, draw=orange!50, right=of step2] (step3) {\textbf{3. Student Responses}\\ \footnotesize 33 Grade 10\\ \footnotesize handwritten scripts};
\node[box, fill=green!10, draw=green!50, right=of step3] (step4) {\textbf{4. Expert Annotation}\\ \footnotesize Double-blind grading\\ $\kappa_w=0.865$};
\node[box, fill=purple!10, draw=purple!50, right=1.2cm of step4] (step5) {\textbf{5. LLM Evaluation}\\ \footnotesize Rubric-based\\ \footnotesize competency prediction};

\draw[arrow] (step1) -- (step2);
\draw[arrow] (step2) -- (step3);
\draw[arrow] (step3) -- (step4);
\draw[arrow] (step4) -- (step5);

\node[draw, dotted, thick, gray, fit=(step1) (step4), inner xsep=0.3cm, inner ysep=0.3cm] (human_box) {};
\node[above=0.1cm of human_box, font=\footnotesize\itshape, text=gray!80!black] {Human Ground Truth Construction};

\end{tikzpicture}
}
\caption{Overview of the benchmark construction and evaluation pipeline.
Human experts design the rubric, create the assessment, and produce
the ground-truth annotations before LLMs are evaluated on the same
rubric-based competency prediction task.}
\label{fig:benchmark_pipeline}
\end{figure}
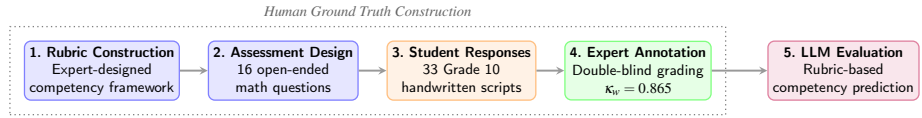

In this study, we benchmark four LLMs with a human-in-the-loop approach to determine which models can best assess students based on the Grade 10 Optional Mathematics Syllabus of Nepal \cite{cdc_optional_math_grade10}. The findings can later be leveraged to develop an autonomous assessment system to assist humans in evaluating students in a more efficient and scientific manner(Fig. \ref{fig:benchmark_pipeline}).

Following are the contributions of the paper:
\begin{enumerate}[noitemsep, topsep=0pt, partopsep=0pt, leftmargin=*]

    \item A competency-aligned Grade 10 Math assessment framework covering topics such as Matrix, Coordinate Geometry, Trigonometry, and Functions
    \item An assessment of cross-competencies that fit overall mathematics levels and topics globally: Comprehension, Knowledge, Operational Fluency, and Behavior and Correlation
    \item A human–LLM comparative evaluation protocol involving two senior mathematics faculty members engaged in competency, rubric, and question design, alongside the competency mapping of students
    \item An empirical analysis across various model scales and architectures.
\end{enumerate}

The rest of this paper is organized as follows: Section 2 discusses Related Works; Section 3 describes the Experiment Design; Section 4 explains the Methodology; Section 5 provides Results and Discussions; Section 6 presents the Conclusions; and finally, Section 7 provides Limitations and Future Recommendations

\section{Related Works}

The integration of Artificial Intelligence (AI) into Competency-Based Education (CBE) has transitioned from theoretical exploration to practical implementation. Recent literature (2023–2025) highlights significant advances in Large Language Models (LLMs) for grading and rubric generation.

\subsection{Global Trends and the Shift to Orchestration}
Research shows an increase in the use of AI for competency evaluation.
Competency-Based Education (CBE) has been widely adopted as a paradigm in which evaluation is heavily dependent on various sources of evidence, such as project portfolios, rubrics, and mentor evaluations, rather than traditional tests \cite{HernandezDeMenendez}. A bibliometric study by \cite{radu2024artificial} shows that scientific output in this area has increased at an average annual rate of 8.43\%, due to the need to scale personalized education. Likewise, in medical education, \cite{do2025trends} found a peak in AI-related publications in 2024, in which there was a shift from basic knowledge validation to the evaluation of sophisticated clinical reasoning skills via Generative AI.

Current models suggest that AI should not only grade but also direct the learning process. \cite{huang2025designing} presented the concept of the ``Adaptive-Autonomy Curve,'' suggesting that AI systems should automatically decrease scaffolding and increase autonomy as competency is verified. At the same time, \cite{nammanee2025ai} presented the AICoLED framework, which employed an Input-Process-Output model to translate digital competencies. The framework used for this study required additional validation because it depended on small expert consensus testing instead of large student evaluation tests. Existing studies mainly investigate higher education and professional fields with limited empirical research on school-level mathematics assessment.

\subsection{Advancements in Automated Assessment Tools}
Technological innovation has directed its efforts toward developing automation solutions for assessment tasks that require extensive manual labor. \cite{wangwiwattana2023automating} proved that Large Language Models (LLMs) could attain a 99.03\% grading accuracy rate compared to human teachers using a structured prompt approach. However, this high accuracy may not translate to secondary education evaluation processes that involve open-ended, complex responses. The SmartRubrics system which \cite{hochstetter2025mapping} described introduced methods that use learning outcomes from syllabus documents to produce assessment matrices.

\subsection{Identified Gaps: The ''Black Box" and Validation Crisis}
Despite these advancements, the reviewed literature identifies three critical limitations hindering the trusted deployment of AI in professional certification:
\begin{enumerate}[noitemsep, topsep=0pt, partopsep=0pt]
    \item \textbf{The ''Black Box" Problem:} \cite{xu2024leveraging} and \cite{faucher2025artificial} point out that the evaluation of AI systems is often a ``black box" problem, in which the underlying logic for a grade is not transparent. This makes it difficult to trust the system, as it may be biased. \cite{fundi2024advancing} further points out that teacher adoption is statistically linked to their confidence in AI ethics; without transparent logic, adoption will be low.
    
    \item \textbf{Technical Limitations in Context:} Although LLMs are highly effective, they are not suited for large contexts. \cite{wangwiwattana2023automating} specifically pointed out that ``longer prompts confused the model,'' resulting in hallucinations or incorrect grades when entire portfolios were graded at once.
    
    \item \textbf{Lack of Empirical Validation:} One of the biggest criticisms that can be found in the scoping review by \cite{do2025trends} is the lack of Randomized Controlled Trials and empirical validation in current research. This is also supported by \cite{hochstetter2025mapping}, who point out that many current systems are prototypes that have not been tested in real-world experimental comparisons to human grading standards\cite{Owan2023}.
\end{enumerate}

\subsection{Bridging the Validation Gap: Positioning This Study in Automated Assessment with Human Benchmarking}
This study directly responds to the validation gap pointed out by \cite{do2025trends} and \cite{xu2024leveraging} by shifting from theoretical prototyping to \textbf{experimental validation}. Although previous studies explored the viability of automation, they rarely provide accurate comparisons with human benchmarks.

To guarantee validity, this study leans towards developing a completely \textbf{automated assessment system} but tests the model's results using \textbf{Human Benchmarking}. Through experiments on actual student work and comparisons between AI-assisted evaluation and expert human evaluators, we determine the ``best model'' for competency validation. This Human-in-the-Loop (HITL) method explains the choice of the model and guarantees that the system is of sufficiently high fidelity for professional education.

\section{Experiment Design}
\subsection{Educational Context and Scope}
The study examines Grade 10 Optional Mathematics curriculum implementation in Nepal which serves as a national requirement for students who plan to study science and technology in higher secondary education. The subject requires students to show their understanding of concepts and their ability to perform tasks because of its high level of difficulty.

Our study focuses on four major topics: Matrix, Coordinate Geometry, Trigonometry, and Functions that require multiple stages of problem resolution and use of mathematical representation and provide opportunities for students to receive partial success marks. The properties of the subject matter enable competency-based evaluation while establishing a real-world environment to assess Large Language Model evaluations against human expert evaluations.

\subsection{Competency Framework Design}

A key contribution of this study is the development of a competency-based assessment framework that evaluates students' \textit{reasoning processes} instead of measuring their answer correctness. The traditional assessment system measures all cognitive functions through one numerical score, which fails to capture essential details about a student's understanding and implementation abilities and their ability to think strategically. We created a competency framework that separates mathematical performance into different components to address this existing limitation.
\paragraph{Cross-Cutting Competencies. } 

Every student answer is assessed on \textbf{four cross-cutting competencies}: \textbf{Comprehension}, \textbf{Knowledge}, \textbf{Operational Fluency}, and \textbf{Behavior and Correlation}. These competencies serve as four distinct analytical tools applied to the same solution. This enables the assessment to make distinctions between whether a student can comprehend the problem, has the conceptual knowledge to solve it, correctly applies mathematical procedures, and appropriately selects or combines solution strategies. This enables two students with the same final solution to have vastly different competency levels.

Each competency is scored on four increasing levels of proficiency: \textit{Awareness}, \textit{Application}, \textit{Mastery}, and \textit{Influence}, which represent increasing levels of depth, autonomy, and integration of reasoning. These levels are not associated with the difficulty of the questions but rather the quality of reasoning in the solution.

\paragraph{Competency Definitions and Observable Evidence.}

Table~\ref{tab:competency_framework} presents the complete competency framework used in this study. The table defines each cross-cutting competency, its corresponding performance levels, and the observable evidence used by both human evaluators and Large Language Models during assessment. This shared operational definition ensures consistency, transparency, and comparability across grading sources.
\begin{table} [htbp]
\centering
\caption{Cross-Cutting Competency Framework with Level-wise Evidence}
\label{tab:competency_framework}
 \scriptsize
\begin{tabularx}{\textwidth}{@{} p{2.8cm} p{2.2cm} X @{}}
\toprule
\textbf{Competency} & \textbf{Level} & \textbf{Observable Evidence} \\
\midrule
\multirow{4}{=}{Comprehension} 
& Awareness & Follows basic directions with limited interpretation. \\
& Application & Understands intent with minimal additional info. \\
& Mastery & Relates prior concepts when solving current problem. \\
& Influence & Translates descriptions into formal math expressions. \\
\midrule
\multirow{4}{=}{Knowledge} 
& Awareness & Possesses prerequisite foundational knowledge. \\
& Application & Demonstrates foundational understanding of the topic. \\
& Mastery & Shows conceptual depth to reason accurately. \\
& Influence & Correlates concepts across different mathematical domains. \\
\midrule
\multirow{4}{=}{Operational Fluency} 
& Awareness & Correctly applies basic rules (e.g., BODMAS). \\
& Application & Accurately solves linear equations with one variable. \\
& Mastery & Handles fractional or multi-step equations. \\
& Influence & Solves higher-order equations with procedural accuracy. \\
\midrule
\multirow{4}{=}{Behavior \& Correlation} 
& Awareness & Identifies the topic or concept of the problem. \\
& Application & Selects an appropriate solution direction. \\
& Mastery & Applies suitable mathematical tools and formulas. \\
& Influence & Integrates multiple concepts for complete solutions. \\
\bottomrule
\end{tabularx}
\end{table}

\paragraph{Rationale and Measurement Protocol.}

The four competencies are specifically structured to be \textbf{independent}, measuring different facets of mathematical thinking. For instance, a student may have strong operational fluency but no conceptual understanding, or have sufficient knowledge but fail to properly integrate concepts for different topics. This allows for a very detailed diagnostic evaluation that cannot be achieved through score-based grading.

Crucially, the system grades \textbf{student reasoning, not answers}. Partial answers, intermediate work, and decision-making are all taken into explicit account, allowing for effective partial credit grading. Comprehension and Operational Fluency are graded throughout the test using competency-specific rubrics. Knowledge is gathered continuously over the course of the test, with earlier questions indicating awareness and later questions indicating a deeper level of application and mastery. Behavior and Correlation are graded mostly through integrative or terminal questions in each topic that require holistic reasoning and concept integration.

This competency framework forms the foundation for the human–LLM benchmarking methodology presented in the following sections.
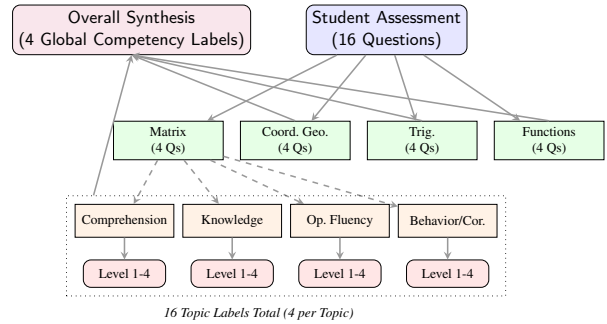
\begin{figure}[htbp]
\centering
\resizebox{0.65\textwidth}{!}{%
 
\begin{tikzpicture}[
   node distance=0.8cm and 0.8cm, 
    every node/.style={font=\sffamily},
    mainnode/.style={draw, rounded corners, fill=blue!10, align=center, minimum width=2.8cm, minimum height=0.9cm},
    topicnode/.style={draw, fill=green!10, align=center, minimum width=2cm, minimum height=0.7cm, font= \scriptsize},
    compnode/.style={draw, fill=orange!10, align=center, minimum width=1.8cm, minimum height=0.6cm, font= \scriptsize},
    levelnode/.style={draw, rounded corners, fill=red!10, align=center, minimum width=1.5cm, minimum height=0.5cm, font=\scriptsize},
    arrow/.style={->, >=stealth, thick, gray!80}
]

\node[mainnode] (exam) {Student Assessment\\(16 Questions)};

\node[topicnode, below left=1.2cm and 1.5cm of exam] (t1) {Matrix\\(4 Qs)};
\node[topicnode, right=0.3cm of t1] (t2) {Coord. Geo.\\(4 Qs)};
\node[topicnode, right=0.3cm of t2] (t3) {Trig.\\(4 Qs)};
\node[topicnode, right=0.3cm of t3] (t4) {Functions\\(4 Qs)};

\draw[arrow] (exam) -- (t1);
\draw[arrow] (exam) -- (t2);
\draw[arrow] (exam) -- (t3);
\draw[arrow] (exam) -- (t4);

\node[compnode, below=0.8cm of t1, xshift=-0.8cm] (c1) {Comprehension};
\node[compnode, right=0.15cm of c1] (c2) {Knowledge};
\node[compnode, right=0.15cm of c2] (c3) {Op. Fluency};
\node[compnode, right=0.15cm of c3] (c4) {Behavior/Cor.};

\draw[arrow, dashed] (t1) -- (c1);
\draw[arrow, dashed] (t1) -- (c2);
\draw[arrow, dashed] (t1) -- (c3);
\draw[arrow, dashed] (t1) -- (c4);

\node[levelnode, below=0.4cm of c1] (l1) {Level 1-4};
\node[levelnode, below=0.4cm of c2] (l2) {Level 1-4};
\node[levelnode, below=0.4cm of c3] (l3) {Level 1-4};
\node[levelnode, below=0.4cm of c4] (l4) {Level 1-4};

\draw[arrow] (c1) -- (l1);
\draw[arrow] (c2) -- (l2);
\draw[arrow] (c3) -- (l3);
\draw[arrow] (c4) -- (l4);

\node[draw, dotted, fit=(c1) (c4) (l1) (l4), inner sep=0.15cm] (fitbox) {};
\node[below=0.1cm of fitbox, xshift=-0.5cm, font=\scriptsize\itshape] {16 Topic Labels Total (4 per Topic)};

\node[mainnode, fill=purple!10, left=1cm of exam] (overall) {Overall Synthesis\\(4 Global Competency Labels)};

\draw[arrow, thick] ([xshift=0.5cm]fitbox.north west) -- (overall.south);
\draw[arrow, thick] (t2.north) -- (overall.south);
\draw[arrow, thick] (t3.north) -- (overall.south);
\draw[arrow, thick] (t4.north) -- (overall.south);

\end{tikzpicture}
}
\caption{Hierarchical competency evaluation rubric showing the mapping from 16 topic-specific labels to four global competency scores.}
\label{fig:evaluation_matrix}
\end{figure}

\subsection{Rubrics Construction and Competency Mapping}

The rubric framework used in this study was designed to make direct connections between assessment items and observable student competencies, offering a framework for both human and LLM assessment. In contrast to general grading rubrics, this rubric addresses \textbf{progressive competency development} on topics and problem types, allowing for smooth assessments of reasoning, knowledge, procedural skills, and conceptual integration.

\paragraph{Design Principles.}

The rubrics were developed with the following principles in mind:

\begin{enumerate}[noitemsep, topsep=0pt, partopsep=0pt]
    \item \textbf{Topic Alignment:} The Grade 10 Optional Mathematics syllabus in Nepal is supported by rubrics which correspond to the specific mathematics units: Matrix, Coordinate Geometry, Functions, and Trigonometry.
    \item \textbf{Competency Coverage:} Each rubric targets one or more different capabilities for each question: \textit{Knowledge, Comprehension, Operational Fluency, and Behavior \& Correlation}.
    \item \textbf{Level-Wise Differentiation:} Each competency is joined across four progressive levels: \textit{Awareness, Application, Mastery, Influence}. An evidence is assigned to each level to denote student performance.
    \item \textbf{Evidence-Based Mapping:} Rubrics make direct references to the question(s) that provide evidence for competency, ensuring traceability and reproducibility.
    \item \textbf{Holistic Integration:} For overall mathematics assessment, rubrics integrate competencies across topics to assess reasoning, procedural skills, and real-world applications.
\end{enumerate}
\paragraph{Rubric Example.}

The table below, Table~\ref{tab:rubrics_example}, shows how the principles are operational in the rubric framework. Each row represents a competency-level pair, and the corresponding observable evidence is directly extracted from the specific questions in the assessment.
\begin{table} [htbp]
\centering
\caption{Rubrics Construction with Competency, Level, and Evidence}
\label{tab:rubrics_example}
 \scriptsize
\begin{tabularx}{\textwidth}{@{} >{\raggedright\arraybackslash}p{2.2cm} >{\raggedright\arraybackslash}p{2.4cm} l >{\raggedright\arraybackslash}X @{}}
\toprule
\textbf{Topic} & \textbf{Competency} & \textbf{Level} & \textbf{Evidence (Question Used)} \\
\midrule
\multirow{4}{*}{Matrix} 
& \multirow{3}{*}{Knowledge} & Awareness & Determine the determinant of a given $2 \times 2$ matrix; find its adjoint \\
& & Application & Identify value of '$k$' making a matrix singular \\
& & Mastery & Solve simultaneous equations using matrix methods \\
\cmidrule{2-4}
& Behaviour \& Correlation & Influence & Explain differences in solutions of two systems with zero determinants \\
\midrule
\multirow{4}{2.2cm}{Coordinate Geometry} 
& \multirow{3}{*}{Knowledge} & Awareness & Determine slope and equation of straight lines \\
& & Application & Find angles between lines; interpret parallelism/perpendicularity \\
& & Mastery & Find equations of lines parallel/perpendicular to a given line \\
\cmidrule{2-4}
& Behaviour \& Correlation & Influence & Find center and radius of circle through three points \\
\midrule
\multirow{4}{*}{Functions} 
& \multirow{3}{*}{Knowledge} & Awareness & Write composition $fg(x)$ and find inverse of $f(x)$ \\
& & Application & Find $(fg)^{-1}(x)$ \\
& & Mastery & Solve functional equations using $gof(x)$ and $f^{-1}(x)$ \\
\cmidrule{2-4}
& Behaviour \& Correlation & Influence & Model investment scenario with companies A and B \\
\midrule
\multirow{4}{*}{Trigonometry} 
& \multirow{3}{*}{Knowledge} & Awareness & Find sine and cosine ratios in right-angled triangles \\
& & Application & Evaluate $\sin(A+B)$, $\cos(A-B)$, $\sin 2A$, $\cos 2A$, $\tan 2A$ \\
& & Mastery & Solve $\cos 3x + \cos x = \cos 2x$ for $0^\circ < x < 90^\circ$ \\
\cmidrule{2-4}
& Behaviour \& Correlation & Influence & Solve real-world height measurement problem with observational error \\
\bottomrule
\end{tabularx}
\end{table}

\paragraph{Project-Specific Implementation.}
This rubric design enables the following:
\begin{itemize}[noitemsep, topsep=0pt, partopsep=0pt]
    \item \textbf{Partial-Credit Scoring:} Students are assessed at the level of competency they have attained in each competency, rather than being scored on a binary scale.
    \item \textbf{Transparent LLM Scoring:} The LLMs are asked questions using the rubric terms and evidence mappings to make competency-level predictions that align with human scoring.
    \item \textbf{Cross-Competency Analysis:} Facilitates analysis of performance across knowledge, comprehension, operational fluency, and behavioral reasoning.
    \item \textbf{Alignment to Pedagogical Objectives:} The rubric highlights both conceptual understanding and application to real-world contexts, aligning with the competency-based objectives of the curriculum.
\end{itemize}
Through the combination of competencies, levels, and evidence, and direct link to the question bank, this rubric design serves as the operational foundation for the following assessment pipeline, ensuring scientific accuracy and reproducibility for both human and LLM scoring(Fig. \ref{fig:evaluation_matrix}).

\subsection{Assessment Tasks and Questions Design}
The assessment tasks were designed to align with the competency framework and capture meaningful evidence of the student's reasoning. For each topic in the Grade 10 Optional Mathematics syllabus, we included a targeted number of questions that are \textit{open-ended,} rather than multiple-choice. Open-ended questions were chosen for several reasons:
\begin{itemize}[noitemsep, topsep=0pt, partopsep=0pt]
    \item \textbf{Capture reasoning steps:} Students are required to show the process involved in arriving at a solution, which enables the evaluation of procedural fluency and strategic thinking.
    \item \textbf{Reveal misconceptions:} Unlike closed-ended questions, open-ended questions reveal partial misconceptions or gaps in conceptual understanding.
    \item \textbf{Allow partial solutions:} Students can show competency even if the final answer is incorrect, which supports error-aware evaluation.
\end{itemize}

Open-ended tasks are therefore essential for competency-based assessment, because they provide rich, observable data across all four cross-cutting competencies (\textit{Comprehension}, \textit{Knowledge}, \textit{Operational Fluency}, \textit{Behavior and Correlation}). This approach guarantees that both human observers and LLMs are able to effectively assess not only what students know but also how they think and apply this knowledge to solve mathematical problems.

\section{Methodology}
\subsection{Dataset Description}
The dataset used for this research was designed to provide a representative sample of student performance in the particular context of the Nepalese Optional Mathematics curriculum. The following are the key features of the data:

\begin{itemize}[noitemsep, topsep=0pt, partopsep=0pt, leftmargin=*]
    \item \textbf{Participant Profile:} A total of 33 Grade 10 students were chosen from a single educational institution. All participants were enrolled in the Optional Mathematics stream.
    \item \textbf{Data Format (Handwritten Responses):} All assessment tasks were completed by hand in order to give a representative sample of student performance in a real-world educational setting.
    \item \textbf{Diversity of Skill Levels:} In coordination with school faculty, students were intentionally selected to represent a diverse range of mathematical proficiencies. In order to accurately assess the competency-based framework across a range of comprehension levels, the selection process sought maximal diversity rather than concentrating on high achievers.
    \item \textbf{Assessment Objective:} The primary goal of the dataset collection was to provide evidence of qualitative competency levels (\textit{Awareness, Application, Mastery, Influence}) rather than to derive simple numerical scores.
\end{itemize}

Before sending for human and LLM assessment, the datasets were preprocessed as follows:
\begin{itemize}[noitemsep, topsep=0pt, partopsep=0pt, leftmargin=*]
    \item The handwritten responses were \textbf{Digitized and Transcribed} with digital scanners and the LLM tool mentioned in Section 3.5.
    \item For complete \textbf{anonymization}, the students' answers were assigned a particular code and the LLMs were also assigned code names rather than their original identifiers.
    \item \textbf{Pages with empty sections or extraneous markings} like crossed answers, doodles, and scribbles were excluded from the scanned copies.
\end{itemize}

\subsection{Human Evaluation and Ground Truth Establishment}

To ensure the reliability and validity of the benchmarking process, a strict human evaluation protocol was followed to set the baseline for the study.

\begin{itemize}[noitemsep, topsep=0pt, partopsep=0pt, leftmargin=*]
    \item \textbf{Expert Evaluators:} Two independent subject matter experts with over 5+ years of experience in the Grade 10 Optional Mathematics curriculum were hired to evaluate the handwritten responses of the students.
    \item \textbf{Blind Grading Protocol:} A double-blind grading protocol was strictly followed. The evaluators were given anonymous student responses and were completely unaware of any Large Language Model (LLM) generated labels or scores. This ensured that the human evaluation was not influenced by any AI-generated scores.
    \item \textbf{Inter-Rater Reliability:} To measure the degree of agreement between the two independent evaluators, the \textbf{Quadratic Weighted Cohen’s Kappa ($\kappa_w$)} coefficient was calculated for all levels of competency.
    \item \textbf{Adjudication and Conflict Resolution:} When the two primary evaluators disagreed on the competency level of a student response, a formal adjudication process was followed. A third senior expert (the Lead Adjudicator) was asked to review the dispute and make a final decision, which was taken as the ground truth for the dataset.
\end{itemize}

\subsection{LLM Selection}
\paragraph{Architectural Implementation and Model Parameters.}The following section describes the technical setup of the multi-model ensemble system used
for the automated assessment of mathematical scripts. To guarantee the experimental integrity and eliminate any random variability, a \textbf{uniform Temperature of 0.1} was set for all models. Additionally, a \textbf{singular Master Evaluation Prompt} was used for all models, ensuring that each model individually assesses the rubric and student responses without any model interaction.

\paragraph{Core Component Analysis.}
The assessment pipeline is based on a multi-model system, where each Large Language Model (LLM) is chosen for a specific operational task based on its performance to latency ratio and reasoning capacity. As shown in Table \ref{tab:model-config}, the system pipeline starts with a multimodal OCR Engine for the accurate transcription of handwritten mathematical scripts.

The core assessment is divided into three distinct levels: EAGLE is responsible for high-speed benchmarking, ORION is responsible for deep reasoning for complex mathematical proofs, and NOVA is a high-throughput control model with a Mixture-of-Experts (MoE) architecture. Finally, LYRA a frontier class model, is the ''Arbiter," which resolves any inconsistencies between the initial model assessments and ensures that the final competency labels are consistent with the predefined human ground truth.
The models were accessed via different deployment platforms, where some APIs were used under paid tiers (e.g., Google Gemini services) and others were accessed through research or free-tier access (e.g., Groq and OpenRouter).

\begin{table}[htbp]
\centering
\caption{Systematic Model Configuration and Deployment Details}
\label{tab:model-config}
\scriptsize
\begin{tabularx}{\textwidth}{@{} l >{\ttfamily\raggedright\arraybackslash}X >{\raggedright\arraybackslash}X >{\raggedright\arraybackslash}X >{\raggedright\arraybackslash}X @{}}
\toprule
\textbf{Component} & \textbf{Exact Model ID} & \textbf{Provider} & \textbf{Primary Role} & \textbf{Architecture} \\
\midrule
OCR Engine & gemini-2.5-flash & Google     & Transcription     & Multimodal \\
EAGLE      & llama-3.1-8b     & Groq       & Latency \& Speed  & 8B Dense   \\
ORION      & llama-3.3-70b    & OpenRouter & Complex Reasoning & 70B Dense  \\
NOVA       & gemini-2.5-flash & Google     & High-Throughput   & Sparse MoE \\
LYRA       & gemini-3-pro     & Google     & Expert Arbiter    & Sparse MoE \\
\bottomrule
\end{tabularx}
\end{table}

\paragraph{Multimodal OCR Engine (The Digitization Layer).}
The \texttt{gemini-2.5-flash} model was selected as the fundamental ingestion layer on account of its \textbf{Native Multimodal} architecture. Unlike conventional OCR techniques, which use post-processing techniques for extracting text, this model treats handwritten examination scripts as input to be visually processed directly. \begin{description}\item[Capabilities:] It demonstrates superior proficiency in deciphering non-standard cursive scripts and preserving the spatial layout of mathematical diagrams.\item[Extended Context:] Its large context window makes it easier to process the current multi-page scripts, thus avoiding data breach in the transcription of long-form proofs.\end{description}

\paragraph{LLM Justification.}
The evaluation setup uses a group of Large Language Models that work together. The ensemble includes analytically oriented dense models (e.g., Orion) and latency-optimized models (e.g., Eagle), enabling controlled architectural comparison. The pipeline culminates in a ''Judicial Arbiter" (Lyra), which utilizes Chain-of-Thought reasoning to handle complex edge cases, ensuring that the final AI-generated labels maintain the exact intent and logical quality of the human expert ground truth while NOVA can handle huge volume.

\subsection{Prompt Strategy}
To ensure evaluation consistency, a singular, identical Master Evaluation Prompt was strictly utilized across all models. This aligned the Large Language Models with the educational objectives of the Grade 10 Optional Mathematics course. As highlighted below, it employed a Role-Based Instruction (RBI) pattern, assigning the model the persona of an ''Expert Mathematics Educator." The prompting strategy was informed by three key considerations:
\begin{itemize}[noitemsep, topsep=0pt, partopsep=0pt]
    \item \textbf{Strict Constraint Enforcement:} The models were strictly prohibited from assigning numerical grades, which helped them to focus more on the qualitative evidence-based competency mapping
    \item \textbf{Multidimensional Output:} The output was organized into six separate sections to ensure that the models evaluated both unit-level performance and intelligent cross-cutting skills (such as Operational Fluency vs. Strategic Correlation).
    \item \textbf{Confidence and Evidence Mapping:} To enhance the transparency of the AI decision-making process, the prompt included a formal ``Evidence Map'' and a confidence score for each competency, addressing the potential black-box problem of hallucinations.
\end{itemize}

\noindent\textbf{Ensemble Roles \& Technical Justification:} 
\begin{itemize}[noitemsep, topsep=0pt, leftmargin=*]
    \item \textbf{ORION (Analytical Benchmark):} 70B parameters provide high logical correctness for multi-step rubrics, preventing ``logical collapse''.
    \item \textbf{EAGLE (Efficiency Baseline):} Speed-optimized baseline (via Groq LPU) at 300 t/s. Tests practicability of lightweight architectures.
    \item \textbf{NOVA (High-Volume Control):} Scalability \& Safety. High-throughput alternative with integrated safety filters for bias mitigation.
    \item \textbf{LYRA (Judicial Arbiter):} Employs Chain-of-Thought (CoT) to resolve disputes between models; aligns with human expert standards.
\end{itemize}
\noindent\textbf{Anatomy of the Competency-Based Assessment Prompt:}
\begin{itemize}[noitemsep, topsep=0pt, leftmargin=*]
    \item \textbf{Persona:} An Expert Mathematics Educator; ensures a neutral, professional, and pedagogically sound tone.
    \item \textbf{Evidence Rule:} No marks or percentages allowed; judgments must be anchored to specific, visible work in the student script.
    \item \textbf{Core Units:} Matrix, Coordinate Geometry, Function, and Trigonometry (Grade 10 Syllabus).
    \item \textbf{Analysis Tiers:} Cross-Cutting Competencies: Comprehension, Knowledge, Operational Fluency, and Behaviour \& Correlation.
    \item \textbf{Output Logic:} Six-section structured report with Markdown tables, Evidence Maps, and an ASCII-based visual summary.
\end{itemize}

\subsection{Level wise and Statistical Rationale}

We employed the Weighted Quadratic Cohen's Kappa ($\kappa_w$) to assess the degree of agreement between the Large Language Model (LLM) and human expert judgment. Because Cohen's Kappa accounts for the ordinal structure of our competency levels (Awareness, Application, Mastery, and Influence), it was selected over accuracy and linear Kappa.

The statistical model is based on the following design considerations:

\begin{itemize}[noitemsep, topsep=0pt, partopsep=0pt]
    \item \textbf{Ordinal Penalty Logic:} Errors are exponentially penalized using a quadratic weighting mechanism ($w_{ij} = (i-j)^2$). For instance, a student's incorrect calculation at the Awareness level when the expert ground truth is at Influence will yield a far lower measure than a minor discrepancy between the neighboring levels.
    \item \textbf{Baseline for Human Reliability: }Before determining the AI performance, we computed the $\kappa_w$ between Teachers 1 (T1) and 2 (T2). ''Almost Perfect" human agreement validates the criteria impartiality. 
    \item \textbf{Multi-Dimensional Variation:} Human vs. Human, Human vs. LLM, and Inter-Model Consensus are the three Kappa statistic tests that we ran. This enables us to determine whether models are adhering to a shared logic that disregards human judgment.
\end{itemize}


\section{Results and Discussions}
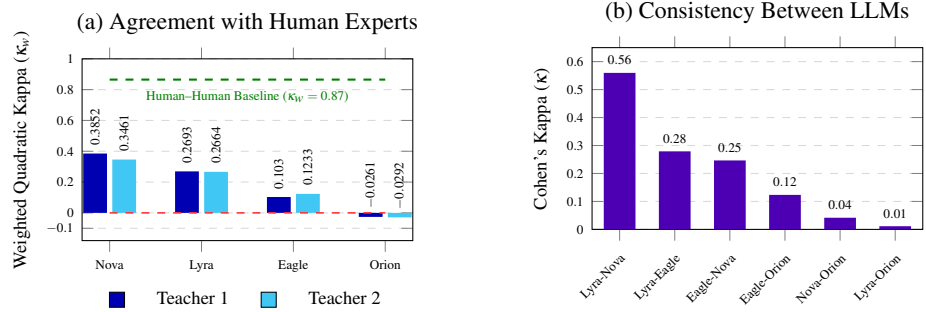
\begin{figure*} [htbp]
\centering
\begin{tikzpicture}
\begin{axis}[
    width=0.49\textwidth,
    height=4.0 cm,
    ybar,
    bar width=9pt,
    ylabel={Weighted Quadratic Kappa ($\kappa_w$)},
    label style={font=\scriptsize}, 
    tick label style={font=\tiny}, 
    symbolic x coords={Nova, Lyra, Eagle, Orion},
    xtick=data,
    ymin=-0.18, ymax=1.00, 
    ytick={-0.1, 0, 0.2, 0.4, 0.6, 0.8, 1.0},
    legend image code/.code={
        \draw[#1, fill] (0cm,-0.1cm) rectangle (0.2cm,0.1cm);
    },
    legend style={
        at={(0.5,-0.22)}, anchor=north, legend columns=-1, column sep=10pt, font= \scriptsize, draw=none
    },
    ymajorgrids=true,
    grid style={dashed, gray!30},
    title={\small (a) Agreement with Human Experts},
    nodes near coords,
    every node near coord/.append style={
        font=\tiny, 
        rotate=90, 
        anchor=west, 
        /pgf/number format/fixed,
        /pgf/number format/precision=4
    }
]
    \addplot[fill=blue!70!black, draw=none] coordinates {
        (Nova,0.3852) (Lyra,0.2693) (Eagle,0.1030) (Orion,-0.0261)
    };
    
    \addplot[fill=cyan!60!white, draw=none] coordinates {
        (Nova,0.3461) (Lyra,0.2664) (Eagle,0.1233) (Orion,-0.0292)
    };

    \draw[dashed, red!80, line width=0.6pt] (axis cs:Nova,0) -- (axis cs:Orion,0);
        
    \draw[dashed, green!50!black, line width=0.8pt] 
        (axis cs:Nova,0.8652) -- (axis cs:Orion,0.8652)
        node[midway, below, font=\tiny, color=green!50!black, yshift=-1pt] {Human--Human Baseline ($\kappa_w=0.87$)};

    \legend{Teacher 1, Teacher 2}
\end{axis}
\end{tikzpicture}
\hfill
\begin{tikzpicture}
\begin{axis}[
    width=0.49\textwidth,
    height=4.0cm,
    ybar,
    bar width=12pt,
    ylabel={Cohen's Kappa ($\kappa$)},
    label style={font=\scriptsize}, 
    tick label style={font=\tiny}, 
    symbolic x coords={Lyra-Nova, Lyra-Eagle, Eagle-Nova, Eagle-Orion, Nova-Orion, Lyra-Orion},
    xtick=data,
    x tick label style={rotate=45, anchor=north east, font=\tiny},
    ymin=0, ymax=0.65,
    ytick={0, 0.1, 0.2, 0.3, 0.4, 0.5, 0.6},
    ymajorgrids=true,
    grid style={dashed, gray!30},
    title={\small (b) Consistency Between LLMs},
    nodes near coords,
    every node near coord/.append style={
        font=\tiny, /pgf/number format/fixed, /pgf/number format/precision=2
    }
]
    \addplot[fill=violet!60!blue, draw=none] coordinates {
        (Lyra-Nova,0.5597)
        (Lyra-Eagle,0.2791)
        (Eagle-Nova,0.2466)
        (Eagle-Orion,0.1234)
        (Nova-Orion,0.0417)
        (Lyra-Orion,0.0114)
    };
\end{axis}
\end{tikzpicture}
\caption{Comparison of LLM performance. Plot (a) shows agreement with human experts, highlighting that while Nova leads, there is a significant gap relative to human-to-human consistency. Plot (b) shows the varying levels of agreement between model pairs.}
\label{fig:results_analysis}
\end{figure*}

The outcomes showed that the LLM ensemble and human experts performed significantly differently(Fig. \ref{fig:results_analysis}). The human-human baseline was almost exactly aligned, while the LLMs had different levels of logical soundness.

\begin{table}[h]
\centering
\caption{Model Agreement Breakdown Across Evaluators.}
\label{tab:model_agreement}

\begin{tabular}{l c c p{4cm}}
\toprule
\textbf{Model} & $\kappa_w$(T1) & $\kappa_w$(T2) & \textbf{Interpretation} \\
\midrule

Nova & 0.3852 & 0.3461 & Stable alignment \\
Lyra & 0.2693 & 0.2664 & Moderate alignment \\
Eagle & 0.1030 & 0.1233 & Low alignment \\
Orion & -0.0261 & -0.0292 & No alignment \\
\bottomrule
\end{tabular}

\end{table}
\footnotetext{T1 and T2 refer to expert mathematics teachers who independently conducted blind grading of student responses under the human evaluation protocol.}

\paragraph{Analysis of Inter-Model Consensus.} 
In addition to human alignment, we evaluated consensus between models (Table \ref{tab:inter_model}). The greatest degree of agreement was found between \textit{Lyra} and \textit{Nova} ($\kappa = 0.5597$), indicating a common internal representation of the Gemini environment. Nonetheless, the Llama and Gemini ecosystems once again reached a consensus of ''Slight," indicating a discrepancy in how the mathematical data should be interpreted.

\begin{table} [htbp]
\centering
\caption{Inter-Model Consensus (LLM vs. LLM)}
\label{tab:inter_model}

\begin{tabular}{l c l}
\toprule
\textbf{Model Pair} & \textbf{Kappa Score} & \textbf{Agreement Level} \\
\midrule

Lyra vs. Nova & 0.5597 & Moderate \\
Lyra vs. Eagle & 0.2791 & Fair \\
Eagle vs. Nova & 0.2466 & Fair \\
Nova vs. Orion & 0.0417 & Slight \\
Eagle vs. Orion & 0.1234 & Slight \\
Lyra vs. Orion & 0.0114 & Slight \\
\bottomrule

\end{tabular}
\end{table}

\paragraph{Discussion.}
The most interesting outcome is the negative Kappa value for \textbf{Orion (70B)}, which indicates ``No Agreement.'' Despite its number of parameters, Orion was unable to perform well under the strict competency mapping criteria, tending to generate assessments in systematic contradiction to human reasoning. 

On the other hand, \textbf{Nova (gemini 2.5 Flash)}, the high-throughput MoE model, was the best-performing observer. This indicates that, in complex rubric-mapped tasks, model ``compliance'' and adherence to instructions in a particular environment might be more important than the total size of model parameters. The fair degree of agreement between Lyra and Nova further supports the existence of a sound, though not perfect, pedagogical reasoning in optimized Sparse MoE models. Furthermore, observative error analysis revealed models occasionally hallucinated, likely by relying on superficial pattern recognition (e.g., matching formulas without verifying procedural execution). This underscores the \textbf{ethical ``Black Box'' risk} of autonomous AI grading, validating our Human-in-the-Loop framework as a necessary safeguard against algorithmic bias. Finally, while tested in Nepal, the evaluated cross-cutting competencies are universally applicable, allowing this framework to scale to \textbf{international contexts}.
\section{Conclusion}
This research work presented a comprehensive assessment of large language models as mathematical competency observers for Grade 10 students, comparing their performance with an ``Almost Perfect'' human expert consensus ($\kappa_w = 0.8652$).

The results show that, although current LLMs are capable of doing competency assessment, they do so to a varying degree of accuracy. The \textbf{Nova (gemini 2.5 Flash)} model was found to be the most practical observer, reaching ``Fair Agreement'' with human observers. The fact that it outperformed a much larger model, the \textbf{Orion (Llama 70B)} model, which experienced a complete logical breakdown ($\kappa_w = -0.0261$), underscores an important paradigm shift: in the context of specialized, rubric-bound assessment, model compliance and follow through are more important than raw model size.

The fact that there is a medium level of inter-model agreement between Lyra and Nova ($\kappa = 0.5597$) indicates that, although current AI assessment is poor, it is not reasonless. This level of agreement gives itself to a \textbf{Human-in-the-Loop} approach. Rather than being decision-makers in their own right, LLMs should be seen as ``first-pass'' observers that point to evidence for human assessment, greatly lessening the workload while maintaining the ``credibility shield'' of human error.

In conclusion, while LLMs are not yet ready to replace human experts in competency mapping, they have already shown a clear future to transform final and growing assessment into a more scalable, evidence-based process.

\section{Limitations and Future Recommendations}
\begin{description}[noitemsep, topsep=0pt, partopsep=0pt]
    \item[Lack of Qualitative Reasoning Analysis:] Because this was a time-bound pilot study, it is based on the \textit{result} of the assessment rather than the \textit{process} by which the AI system arrives at its reasoning. We did not conduct a qualitative audit of the internal models. A given LLM could correctly identify a competency level based on superficial pattern recognition rather than a deep pedagogical understanding.
    
    \item[The Orion Scaling Paradox:] The ``No Agreement'' outcome of the Orion (70B) model is an anomaly, indicating highly parameterized dense models are potentially more vulnerable to ``instruction drift'' when presented with multi-dimensional rubrics than optimized sparse MoE models.
    
    \item[Sample Size and Scope:] The current sample size of 33 students limits generalization and may not reflect the entire range of mathematical misconceptions in larger populations.
    
    \item[OCR and Transcription Bias:] While using Gemini 2.5 Flash with a targeted multimodal prompt minimized traditional OCR issues, minor transcription bias may still exist.
\end{description}

\subsection{Future Recommendations}
To bridge the existing gap between ``Fair Agreement'' and the level of reliability required for autonomous classroom implementation, the following research directions are recommended:

\noindent \textbf{Verifying the AI’s Logic (Reasoning Audits).} Future studies must include expert qualitative reviews of the AI’s inference explanations to verify true pedagogical alignment, ensuring it is not simply performing pattern recognition.
\noindent \textbf{Dataset Expansion.} Conduct large-scale studies across multiple institutions to ensure broader generalizability. 
\noindent \textbf{Deep Analysis of the ``Scaling Paradox''.} Empirically investigate the mechanisms of ``instruction drift'' causing large dense models to fail under complex rubrics.
\noindent \textbf{Implementing a ``Second Opinion'' (The Critic Loop).} To avoid the problems of biased or overly generous grading, we recommend a two-AI system. In this system, one AI is the \textit{evaluator} who assigns the grade, while another \textit{critic} AI looks for evidence that the first AI has missed.
\noindent \textbf{Topic-Specific Calibration.} The requirements of grading differ depending on the mathematical topic. For instance, grading on \textit{trigonometry} requires an evaluation of visual-spatial skills.
\noindent \textbf{Consistency and Stability Over Time.} A good grading tool should provide consistent results. We recommend that these models be tested over a full academic year to ensure that they provide consistent progress tracking.

This study demonstrates that large language models can function as complex \textbf{assistive} technologies for competency-based assessment. While they are capable of recognizing patterns and producing reasonably accurate competency evaluations, human expertise remains essential for ensuring validity and reliability in educational assessment.

\section*{Acknowledgements}
The authors thank Occidental Public School and its students for providing the opportunity to collect the data. They also thank Mrs. Rojina Shrestha and Mrs. Kabita Shrestha of Sunway College for their assistance in data pre-processing.

\section*{Disclosure of Interests}
The authors have no competing interests to declare.

%
%
\begingroup
\scriptsize
\bibliographystyle{splncs04}
\bibliography{references}
\endgroup
\end{document}